\documentclass[10pt,conference]{IEEEtran}
\IEEEoverridecommandlockouts
\usepackage{cite}
\usepackage{subfigure}
\usepackage{array}
\usepackage{amsmath,amssymb,amsfonts}
\usepackage{algorithmic}
\usepackage{graphicx}
\usepackage{textcomp}
\usepackage{url}
\usepackage{fixltx2e}
\usepackage{xcolor}
\usepackage{csquotes}
\usepackage[bookmarks=false]{hyperref}
\usepackage{multirow}
\usepackage{makecell}

\graphicspath{ {images/} }

\def\BibTeX{{\rm B\kern-.05em{\sc i\kern-.025em b}\kern-.08em
    T\kern-.1667em\lower.7ex\hbox{E}\kern-.125emX}}
\begin{document}

\title{A real-time iterative machine learning approach for temperature profile prediction in additive manufacturing processes \\
}

\author{
    \IEEEauthorblockN{Arindam Paul\IEEEauthorrefmark{1}, Mojtaba Mozaffar\IEEEauthorrefmark{2},
    	Zijiang Yang\IEEEauthorrefmark{1},
    	Wei-keng Liao\IEEEauthorrefmark{1}, 
    	Alok Choudhary\IEEEauthorrefmark{1}, 
    Jian Cao\IEEEauthorrefmark{2}, Ankit Agrawal\IEEEauthorrefmark{1}}
    \IEEEauthorblockA{\IEEEauthorrefmark{1}Department of Electrical and Computer Engineering
    \\\{arindam.paul, zyz293, wkliao, choudhar, ankitag\}@eecs.northwestern.edu}
    \IEEEauthorblockA{\IEEEauthorrefmark{2}Department of Mechanical Engineering\\
   mojtabamozaffar2020@u.northwestern.edu, jcao@northwestern.edu
}}

\maketitle

\begin{abstract}
Additive Manufacturing (AM) is a manufacturing paradigm that builds three-dimensional objects from a computer-aided design model by successively adding material layer by layer. AM has become very popular in the past decade due to its utility for fast prototyping such as 3D printing as well as manufacturing functional parts with complex geometries using processes such as laser metal deposition that would be difficult to create using traditional machining. As the process for creating an intricate part for an expensive metal such as Titanium is prohibitive with respect to cost,  computational models are used to simulate the behavior of AM processes before the experimental run. However, as the simulations are computationally costly and time-consuming for predicting multiscale multi-physics phenomena in AM,  physics-informed data-driven machine-learning systems for predicting the behavior of AM processes are immensely beneficial. Such models accelerate not only multiscale simulation tools but also empower real-time control systems using in-situ data. In this paper, we design and develop essential components of a scientific framework for developing a data-driven model-based real-time control system. Finite element methods are employed for solving time-dependent heat equations and developing the database. The proposed framework uses extremely randomized trees - an ensemble
of bagged decision trees as the regression algorithm iteratively using temperatures of prior voxels and laser information as inputs to predict temperatures of subsequent voxels. The models achieve mean absolute percentage errors below 1\% for predicting temperature profiles for AM processes. The code is made available for the research community at https://github.com/paularindam/ml-iter-additive.

\end{abstract}


\section{Introduction}
	\label{sec:intro}
Additive Manufacturing (AM) is a modern manufacturing approach in which digital 3D design data is used to build parts by sequentially depositing layers of materials~\cite{watson2018decision}. AM techniques are becoming very popular compared to traditional approaches because of their success in building complicated designs, fast prototyping, and low-volume or one-of-a-kind productions across many industries. Direct Metal Deposition (DMD)~\cite{ding2016development} is an AM technology where various materials such as steel or Titanium are used to develop the finished product. Computational simulations are an essential part of the AM design and optimization as they eliminate the trial and error on expensive manufacturing processes. Finite element-based multi-physics simulation models (FEM)~\cite{ding2011thermo, yan2018data} are designed to replicate the AM process before generating the required part using AM. However, FEM-based simulations are computationally costly and time-consuming. This leads to the motivation to develop a predictive tool based on machine learning (ML) that can instantaneously yield the simulation result instead of performing expensive physics-based simulations. 

A real-time AM control system can be useful in manufacturing because it can control machines considering the changes in the environment and the machine itself. This can be more important in AM since most of the vital parameters in the quality of final product change considerably during the build process. The temperature field created while building a part using AM is one of the critical components in determining microstructure, porosity, and grain size. This system requires a fast data-driven predictive model that can relate machine parameters and replicate desired property behavior accurately using ML techniques, without the need for computationally expensive calculations.  There has been an upsurge of interest in the manufacturing community to connect and share data between geographically distributed facilities~\cite{correa2018new, garcia2019sustainable}. We believe a significant amount of experimental data will be available in the near future for manufacturing processes, especially AM. This urges the scientific community to develop suitable data-driven tools and techniques. 

In this work, we use Generalized Analysis for Multiscale Multi-Physics Application (GAMMA)~\cite{smith2016thermodynamically, mozaffar2019acceleration}, a FEM based method for developing the database to train our model-based control system. GAMMA is used to solve the time-dependent heat equation and simulate the manufacturing DMD process at the part scale. As the AM process is a spatiotemporal phenomenon (since there is cooling and reheating depending on whether and when a neighboring element is created), any approach for predicting the temperature profile must include the information about neighboring voxels as well as temporal information. In our proposed approach, we harness this characteristic of the AM process during feature reconstruction for our learning system. The input features for our proposed system include the distance of a given voxel from the current laser beam in the $x$, $y$ and $z$ axes, laser intensity, time at which the point is created, the time elapsed, and tool speed. One of the advantages of a real-time system is instead of training a prior model ahead of time, one can be developed in-situ. This is crucial for the versatility of ML-driven control system, especially as factors such as laser path, laser speed, and laser temperature can largely influence the temperature profile in AM processes which in turn can predict presence of residual stress~\cite{li2018residual}. Residual stress caused in AM is the critical issue for fabricated metal parts since steep residual stress gradients generate distortion which dramatically deteriorate the functionality of the parts. 

 The proposed approach uses extremely randomized trees (ERTs)~\cite{geurts2006extremely}, a tree-based ensemble algorithm to iteratively train a model-based control system. A model is developed on the features of first $m$ voxels to predict the temperature of next $n$ voxels at the first stage, and then iteratively a new model is developed at every subsequent stage using the ground-truth temperature of $m$ voxels as well as the predicted temperature of the $n$ voxels.  The result of this work is a real-time iterative supervised predictive model that achieves \% mean absolute error (\% MAE) below 1\% for predicting temperature profiles for AM processes. The iterative model outperforms a traditional model that does not use predicted intermediate voxel temperatures. The code is made available for the research community at \url{https://github.com/paularindam/ml-iter-additive}. 

The rest of the paper is organized as follows. Section~\ref{sec:background} provides a brief background of AM and DMD, and the FEM code used for developing the database and some related works for application of machine learning in materials informatics, and specifically AM. In Section~\ref{sec:data}, we explain the generation and transformation of the dataset and describe the input features and voxel categories. We describe the motivation and methodology and development of the dataset in Section~\ref{sec:method}. We discuss the experimental settings and results in Section~\ref{sec:results}, and finally in Section~\ref{sec:conclusion}, we summarize our conclusions with some future directions.

\section{Background and Related Works}
	\label{sec:background}
In this section, we present a background of AM and DMD, and the FEM code used for developing the database and some related works to the application of machine learning in materials informatics. 

\subsection{Additive Manufacturing: Overview}

The initial development process for creating a three-dimensional
object using computer-aided design (CAD) for a layer by layer deposition was realized due to a desire for rapid prototyping~\cite{frazier2014metal,wong2012review}. It reduced the time-cycle of realizing an initial prototype after the conception of design by engineers~\cite{ngo2018additive}. Among the major advances that this process presented to product development are the time and cost reduction, and the shortening of the product development cycle. Further, it led to the possibility of creating shapes that were difficult to be machined using traditional manufacturing processes. 

AM can appreciably reduce material waste, decrease the amount of inventory, and reduce the number of distinct parts needed for an assembly~\cite{mueller2012additive,kruth1998progress}. Further, AM can reduce the number of steps in a production process, both in the case of tool making as well as direct manufacturing, reducing the need for manual assembly~\cite{faludi2015comparing}. Besides, AM processes can significantly reduce the total amount of tooling required and its impact on the cost~\cite{matilainen2014characterization}. AM parts can be manufactured in an almost final state, thus reducing the amount of connecting parts required to put them together and decreasing part count~\cite{huang2013additive}. 

\subsection{Direct Metal Deposition}
\begin{figure}[!ht]
\centering
\subfigure[DMD overall setup]{%
\includegraphics[width=.4\textwidth]{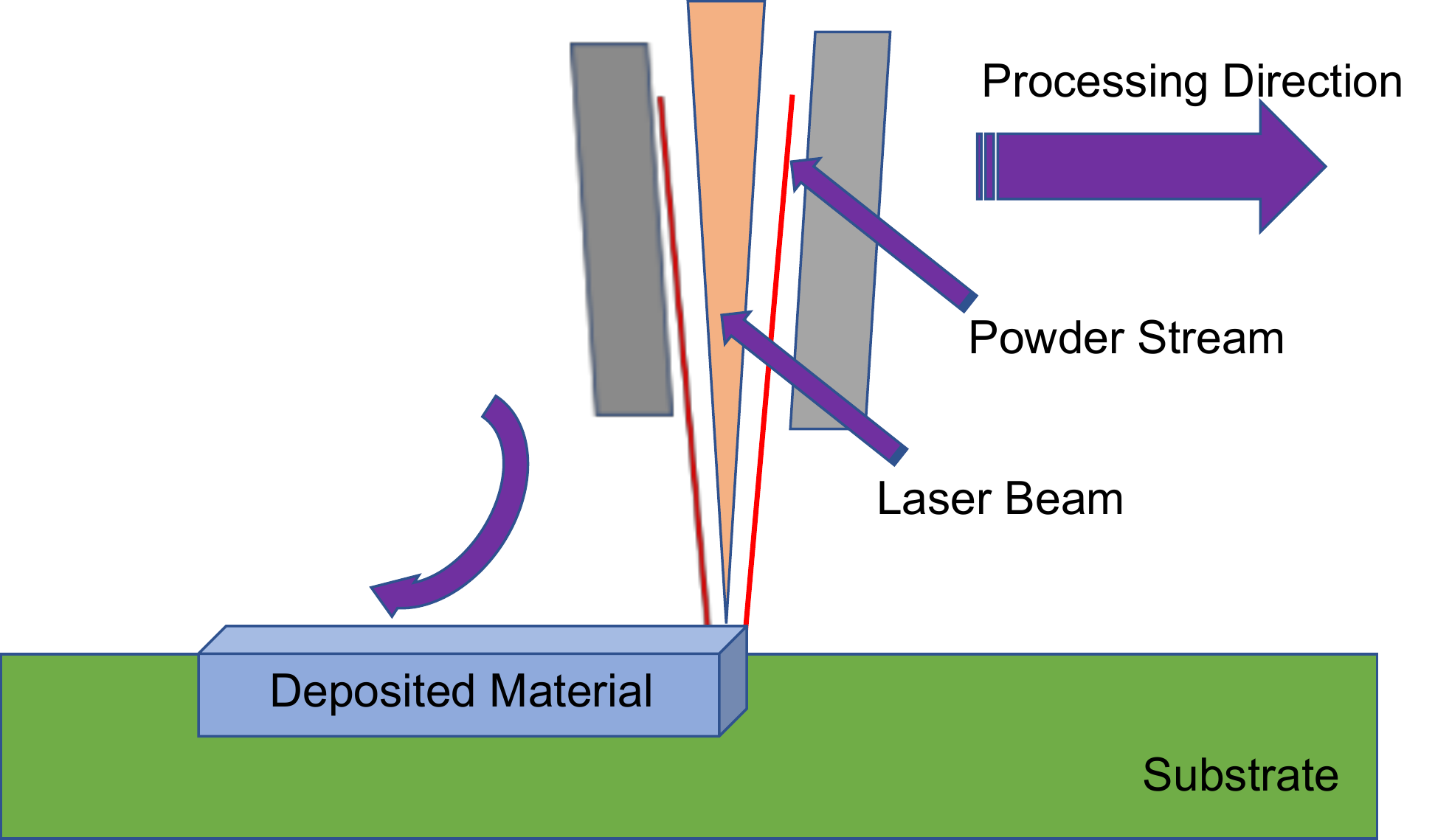}
}
\quad
\subfigure[DMD Laser Path on Substrate ]{%
\includegraphics[width=.4\textwidth]{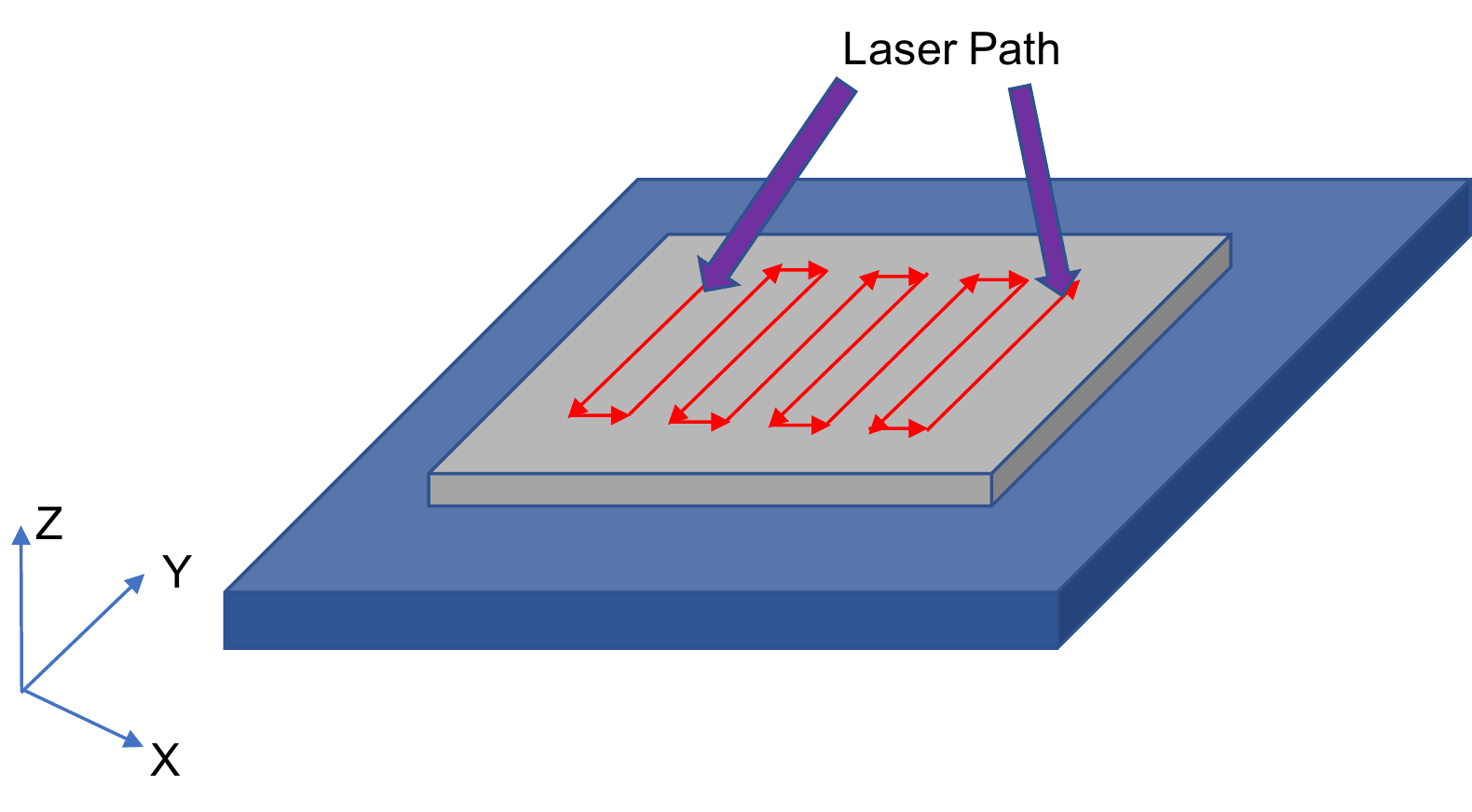}
}
\caption{Additive Manufacturing using Direct Metal Deposition (DMD) process. The laser source provides the heat while the powder stream provides the metal for the deposition. The metal powder gets melted by the heat from the laser beam and deposited on the substrate. The laser scans over the substrate in a zigzag motion.}
\label{fig:DMD}
\end{figure}

\begin{figure}[]
\centering
\includegraphics[width=.4\textwidth]{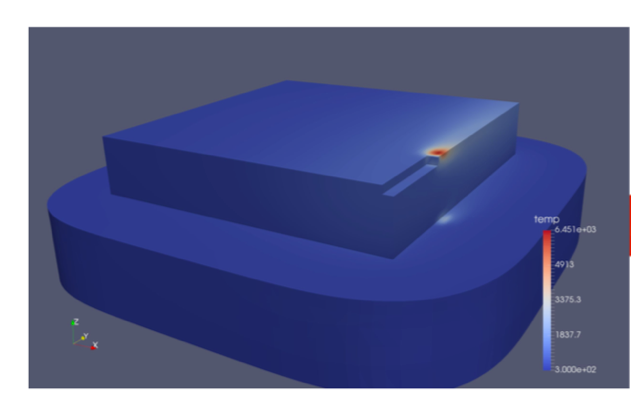}
\caption{The simulated metal surface built using DMD is depicted in the figures. The first figure demonstrates the metal created using DMD on the substrate with the temperature scale. The color of the metal surface indicates the spatio-temporal characteristic of the DMD process.}
\label{fig:additive}
\end{figure}

DMD~\cite{peyre2008analytical} is an additive manufacturing technology using a laser to melt metallic powder.  DMD processes can produce fully dense, functional metal parts directly from CAD data by depositing metal powders using laser melting and a patented closed-loop control system to maintain dimensional accuracy and material integrity~\cite{qi2006numerical}.  Heat is generated as a focused heat source such as a laser to sufficiently melt the surface of the substrate and creates a melt pool. A  focused powder stream provides material for the melt pool using to form a raised portion of the material. The nozzle is moved over the substrate using a computer-controlled positioning system to create the desired geometry. This is illustrated in Figures~\ref{fig:DMD} and ~\ref{fig:additive} that depict the DMD process and laser motion, and the metal surface built across layers, respectively. 

\subsection{Finite Element Method Solver}
Finite element method (FEM) analysis is a numerical approach for solving differential equations over complex geometries with broad applications in simulating structural properties and fluid dynamics~\cite{fish2007first}. In this method, first the domain is discretized into small elements, and then a system of equations is assembled over all the elements. GAMMA is a FEM framework that solves transient heat transfer equations for metal powder-based AM processes such as Directed Energy Deposition (DED)~\cite{stender2018thermal} and Selective Laser Melting (SLM)~\cite{smith2016linking}. Although an accurate thermal analysis of AM provides vital information for determining microstructure evolution and mechanical performance of the part~\cite{wolff2017framework,yan2018data}, this kind of analysis can take weeks or months of computing time and therefore too computationally expensive for large-scale problems or optimization purposes~\cite{francois2017modeling}. For a given set of processing parameter inputs such as build geometry, laser power, and scan speed, GAMMA calculates spatially-dependent thermal histories within the part, such as temperature profiles and maximum cooling rate. In this work, we use GAMMA to generate the database to train our ensemble model.

\subsection{Related Work}
The idle pace of development and deployment of new/improved materials has been deemed as the main bottleneck in the innovation cycles of most emerging technologies~\cite{kalidindi2015data}. Exploring and harnessing the association between processing, structure, properties, and performance is a critical aspect of new materials exploration~\cite{olson1997computational, agrawal2019deep, sundararaghavan2007linear, cang2017microstructure}.  Data-driven techniques provide faster methods to know the important properties of materials and to predict feasibility to synthesize materials experimentally. This can expedite the research process for new materials development. Many initiatives to computationally assist materials discovery using ML techniques have been undertaken~ \cite{paul2019opv, kusne2014fly, paul2019transfer, jha2019irnet, jung2019efficient, jha2018elemnet, rahnama2018machine, paul2018chemixnet, pozun2012optimizing, paul2018sampling, ramprasad2017npjreview, paul2019feedback, jha2019padnet}. 

There has been some limited work on the application of ML techniques for AM processes. Mozaffar et al.~\cite{mozaffar2018data} proposed a data-driven approach to predict the thermal behavior in a directed energy deposition process for various geometries using recurrent neural networks. The proposed approach mapped the position of a point on the printing surface, the time of deposition, the distance of the closest cooling surface, and laser parameters with the thermal output. Baturynska et al.~\cite{baturynska2018optimization} propounded a conceptual framework for combining FEM and ML methods for optimization of process parameters for powder bed fusion AM. Choy et al.~\cite{choy20163d} designed a novel recurrent neural network architecture 3D recurrent reconstruction neural network (3D-R2N2) that learned mapping from images of objects to their underlying shapes in an AM simulation environment. Scime et al.~\cite{scime2018anomaly, scime2018multi, scime2019using} developed supervised as well as unsupervised models for detecting irregularities and flaws on the laser bed during the AM process. 

\section{Data}
	\label{sec:data}
In this section, we explore the generation of the FEM dataset, the transformation of the FEM dataset for machine learning and description of input features and voxel categories. 

\subsection{Data Generation}
The database for training the supervised model was developed using GAMMA by solving time-dependent heat equations and simulating the manufacturing process at the part scale. It provides temperature and heat flux for every time step for every element that is created during the AM process. In this work, we utilize a GAMMA FEM simulation of 20 mm x 20 mm x 3 mm cuboidal dimensions. A mesh voxel size of the edge length of 0.5 mm was used. This refers to a cross-section of 40 x 40 voxels along the $x$ (lateral) and $y$ (longitudinal) axes, and  Therefore, there are 40 x 40 x 6 voxels in the simulation or 9600 voxels. The time taken for the FEM simulation is about an hour. 

The voxel edge length of 0.5 mm chosen in this work is fairly coarse. However, the time taken for a simulation exponentially increases as the mesh voxel size is made finer. For instance, if we reduce the mesh size to half or 0.25 mm, the FEM simulation would take 9 hours. Moreover, the number of data points is in the order of O($n^2$) in terms of the voxels. This is because the FEM simulation contains the temperature history of a voxel from the time of creation to the end of the simulation. Therefore, if one voxel is created at each timestep, there will be $n$ data points pertaining to the first voxel created, $n-1$ data points for the second voxels and so on resulting in $n*(n+1)/2$ data points. However, as the laser deposition process creates multiple voxels at the same time-instant, the number of total data points is significantly smaller but nonetheless of the order of O($n^2$). This is because each data point corresponds to a unique ($x,y,z,t$) where ($x,y,z$) represents an individual voxel and $t$ represents the timestep. In this case, for the 9600 voxels, there are about 9.051652e+06 or about 9.05 million data points. It must be noted that each timestep does not create the same number of voxels as the simulation mimics the weaving (zigzag) motion of the laser (illustrated in Figure~\ref{fig:DMD}(b)). More voxels are created during the lateral movements as compared to when the laser motion reverses.
 
  We chose this simulation size by making a trade-off between a very large simulation that would take days or weeks and potentially create trillions of data points and a small simulation that have too few data points to train and evaluate the proposed approach rigorously. 
    
\subsection{Data Pre-processing}
\begin{figure}[]
\centering

\subfigure[Temperature Profile of overall Additive Manufacturing Process at the end of the FEM simulation]{%
\includegraphics[width=.5\textwidth]{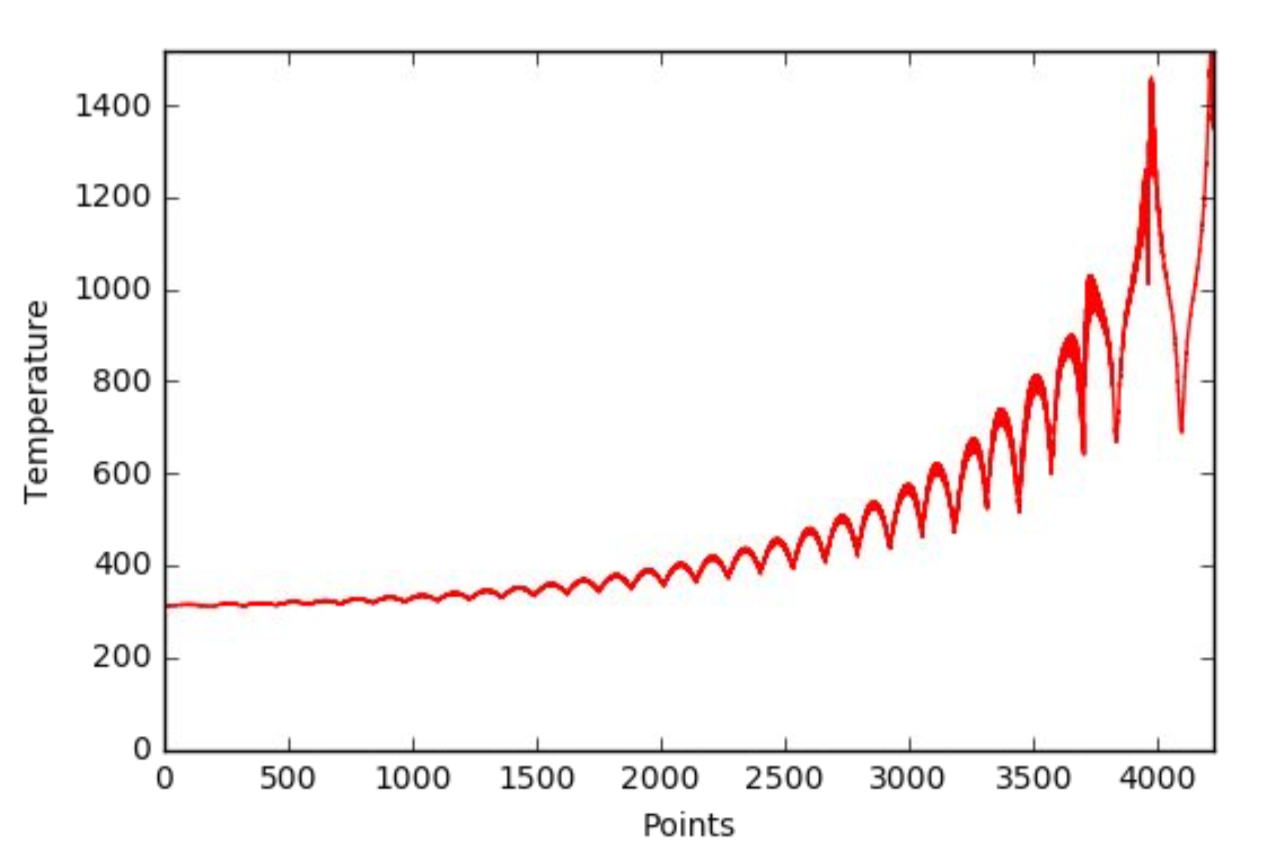}
}
\quad
\subfigure[Temperature Profile across different laser intensities and layers]{%
\includegraphics[width=.5\textwidth]{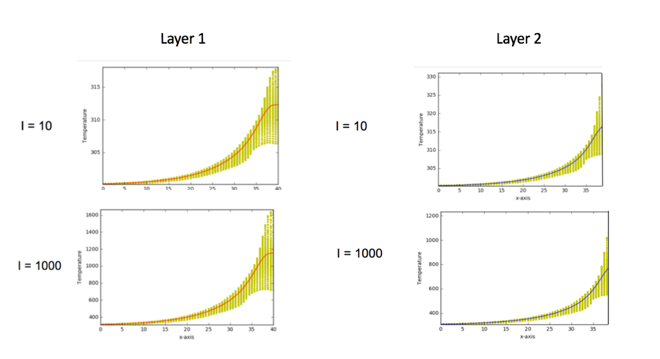}
}
\caption{Temperature profiles for the DMD process. The temperatures are in Kelvin (K) scale. }
\label{fig:tempprofile}
\end{figure}

Figure~\ref{fig:tempprofile}(a) illustrates the overall temperature profile for the DMD process at the end of the AM process. The index of the point in the x-axis demonstrates the time of the creation of the point. We can observe that the points with the lower index or those created earlier slowly approach the room temperature. However, the temperature of the points created later is much higher. Although the overall temperature curve goes higher, we can observe troughs and crests. The troughs are a result of slow cooling of a point created by DMD, and the crest happens when a nearby voxel gets created or heated up. Figure~\ref{fig:tempprofile}(b) illustrates the temperature pattern across different layers are similar, as well as across different laser intensities. Therefore, for a higher laser intensity, we can observe a steeper curve. The temperature curves indicate that the AM temperature profile has spatiotemporal as well as other factors dependent on the laser. 

There are many features that impact the temperature of a given voxel. The most important elements are the position of the voxel ($x$,$y$,$z$) and the time elapsed after the creation of a voxel. Instead of considering the absolute voxel ($x$,$y$,$z$) position, we consider the distance in the $x$,$y$,$z$ with the current position of the laser. The temperature of a given voxel change with time: cooling or heating. As time passes, the temperature of a given voxel reduces. However, if a new voxel is created proximal to the given voxel, this leads to the increase of the temperature of the voxel. However, the temperature profile fluctuates because of cooling and subsequent reheating due to new material creation. Hence, the feature set for training the supervised model that is agnostic of the temperature of adjacent elements would not provide sufficient information for a supervised learning algorithm to learn the AM process. The temperature of each element is influenced by the temperature of its neighboring elements. The following are the input features used for building the proposed predictive model: 

\begin{itemize}
\item Historical Features: Temperature of the given voxel at $t-1$ through $t-5$ (if applicable)
\item Spatio-Temporal Features: Temperature of neighboring 26 voxels at $t-1$
\item Spatial Features: relative $x$, $y$ and $z$ coordinates of the current voxel with respect to the current position of the laser 
\item Temporal Features: Time of voxel creation and time elapsed since the creation of given voxel 
\end{itemize}

\begin{figure}[]
\centering
\includegraphics[width=.35\textwidth]{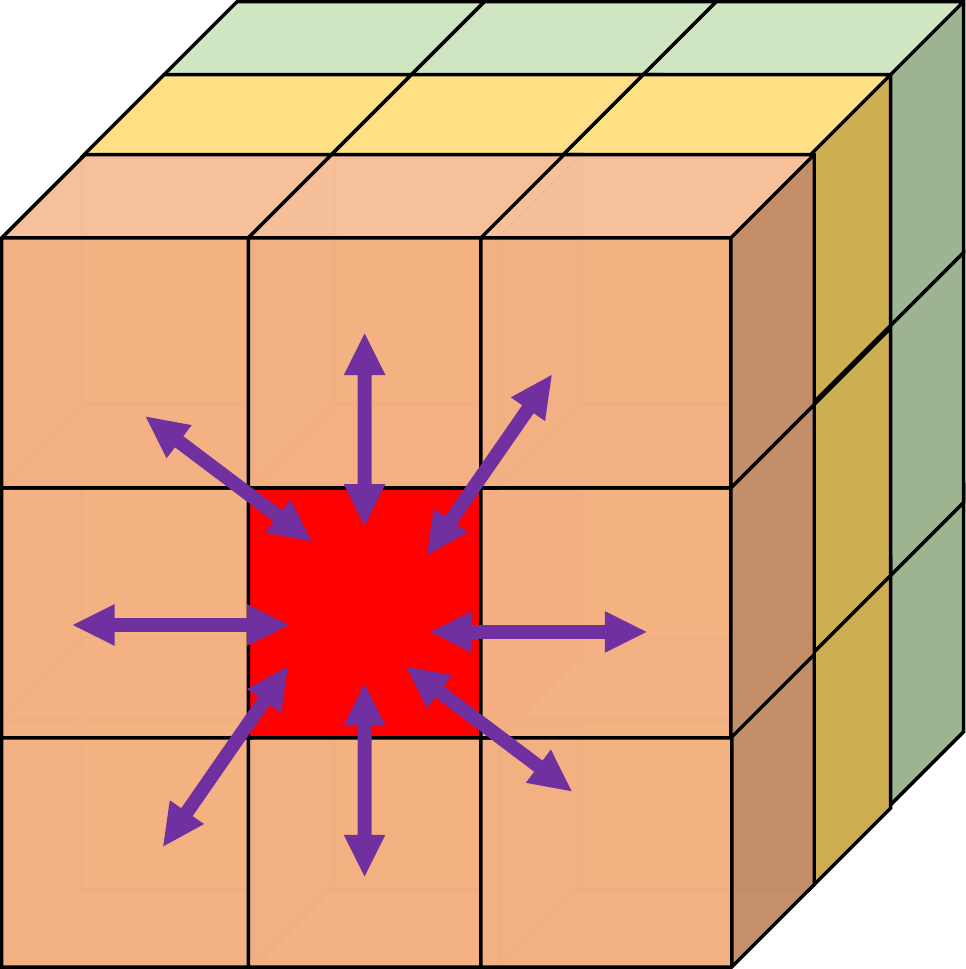}
\caption{Illustration of the cross-section of the AM-surface to represent conduction of heat on target voxel (labeled in red) from neighboring voxels. However, as this is a 2D cross-section of a voxel, there are eight neighboring voxels indicated by arrowheads. In three dimensions, a voxel is surrounded by 26 neighboring voxels. The different colors of adjacent layers indicate the relative temperature. Layers farther away from a newly created voxel are comparatively cooler: green indicating cool, yellow indicating warm and orange indicating hot. }
\label{fig:voxelHeat}
\end{figure}

It is to be noted that the current position of the laser is dependent on both the tool path as well as the tool speed of the laser.  Further, it is not necessary that all the input features are available for all the data points. This is possible in case of voxels at the edge that does not have neighboring voxels or the absence of temperature history of the given voxel. If the temperature of any feature is missing, we assign a dummy value of -99 as most machine learning algorithms do not accept missing values. One of the essential elements of selecting features is selecting independent attributes. We attempt to build a predictive model which only depends on elements which can be reproducible independent of the dataset on which it has been trained. Figure~\ref{fig:voxelHeat} depicts the cross-section of the AM-surface to represent conduction of heat between neighboring voxels.

 \subsection{Voxel Categories}

 We classify the voxels into five categories based on the spatial location of the voxel. As the temperature profiles of voxels surrounded by other voxels may differ from voxels at the periphery, we wanted to investigate if the voxels at the outer edge that have one or more missing neighboring voxels are predicted worse than the interior voxels. This is because our proposed model is dependent on the temperature of the neighboring voxels. To characterize this, we categorize the voxels into five categories. 
\begin{itemize}
\item Interior: all neighboring voxels present 
\item Edge (Lateral): neighboring voxel on the x-axis missing
\item  Edge (Longitudinal): neighboring voxel on the y-axis missing
\item Edge (Vertical): neighboring voxel on the z-axis missing 
\item Edge (Diagonal): a neighboring voxel on the planar or cubical diagonal is missing (but no lateral, longitudinal or vertical neighbors are missing)
\end{itemize}

To avoid confusion, we avoid categorizing a single voxel into multiple categories. If a voxel has a missing neighbor on the $x$-axis, it is considered an edge (lateral) voxel even if it has a missing $y$ or $z$-axis neighbor. Similarly, if a voxel has a missing neighbor on $y$-axis but no missing edge on the $x$-axis, it is considered as an edge (longitudinal) even if there is a missing $z$-axis neighbor. We decide in this fashion as we can anticipate that newly created voxels might have a missing voxel vertically above ($z$ axis). Therefore, a voxel that has $x$ or $y$-axis neighbors missing are considered more distinct than a missing $z$-axis neighbor. If a voxel has any neighbor missing apart from the immediate adjacent neighbor along the $x$, $y$ and $z$ axes, it is considered a diagonal edge voxel. It is noteworthy that when we categorize a voxel, we do it at a specific time $t$. This is because for a given newly created voxel at layer $l$ would be an edge(lateral) voxel at the time of creation, but the layer $l+1$ is deposited on top of this voxel, it would be an interior voxel. 

\section{Method}
	\label{sec:method}
\begin{figure}[]
\centering
\includegraphics[width=.35\textwidth]{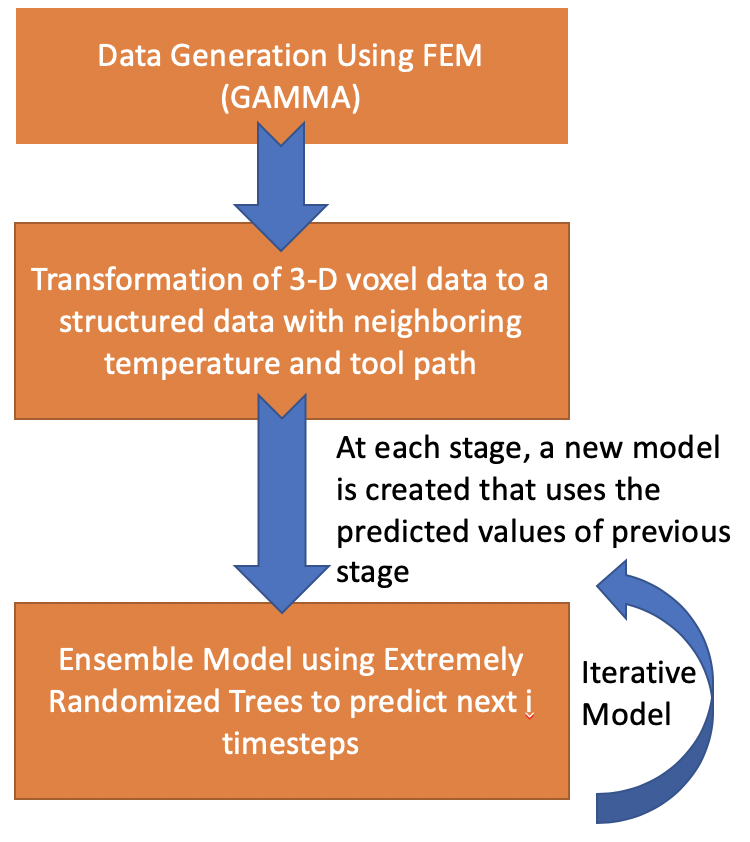}
\caption{The overall methodology of the proposed multi-stage iterative model for predicting temperature profile of an additive process}
\label{fig:iterativeFlow}
\end{figure}

The motivation and methodology of the proposed iterative approach are outlined in this section. Figure~\ref{fig:iterativeFlow} illustrates the flow diagram of the proposed methodology. 

\subsection{Motivation for real-time system}
Control systems in manufacturing can be divided into two broad categories~\cite{tapia2014review}. The first class is error-based control systems in which changing parameters (parameters of manufacturing machine such as laser power, speed) are estimated and based on the error values from the experiment, the initial guess is corrected until the desired criteria is met. The second class is model-based in which instead of estimating the initial value of machine parameters, they will be determined by a model. 

While an error based control system can be useful in many applications such as motion control, its application in AM process parameter control is not common because a significant deviation will ruin the part. Developing control manufacturing processes in a way to achieve desired properties in the final product is not a new attempt. It started from simple trial and errors and gradually developed to complicated multiple-layer feedback control systems to manipulate system settings for real-time control. However, growing demand for controlling more and more detailed and complicated properties of products overpassed current science and many scientists tried to come up with new methods to overcome this challenge. As a data-driven methodology is more intuitive with a model-based system, our proposed approach outlines such a control system where the model is developed by training a machine learning algorithm.

\subsection{Iterative Ensemble Model}
We explored across many regression algorithms for the developing our models including linear regression (ordinary least square), regularized linear regression: Lasso (L1-regularization) and Ridge (L2-regularization), boosted and bagged decision trees. We did not consider neural networks for this framework. Although, a recurrent neural network model trained on temporal features can be combined with a feed-forward neural network trained on non-temporal features, training deep neural networks would take hours to train which is many order of magnitudes time more than the simulation time for FEMs and not feasible for a real-time prediction system where training has happened in-situ. Further, algorithms based on autoregression and moving average such as ARIMA~\cite{ulrich2005timeseries} would not be able to capture spatial non-temporal relationships. This is also evident from our benchmarking experiment in Table~\ref{tab:estimators_compare}.  We considered two metrics $R^2$ (coefficient of determination) and \% MAE  to evaluate the performance of the models. 

 \begin{table}[]
\centering
\small
\caption{Comparison of performance for different machine learning algorithms with corresponding $R^2$ and \% MAE based on training on the first 200 timesteps and predicting next 300 timesteps. For each algorithm, we explore various hyperparameters and present the best model. }
\label{tab:estimators_compare}
\begin{tabular}{|l|c|c|c|}
\hline
\textbf{Algorithm} & \textbf{$R^2$} & \textbf{\% MAE} & \textbf{Training Time}\\ 
  &  &  & \textbf{(in seconds)}\\ \hline
Linear Regression &0.23&25.08&0.52 \\ \hline
Lasso Regression &0.21&23.11&0.53 \\ \hline
Ridge Regression &0.38&17.28&0.56 \\ \hline
ARIMA&0.15&29.39&0.67 \\ \hline
Decision Trees & 0.76&9.74&2.30 \\ \hline
AdaBoost (20 trees) & 0.89&9.40&9.89\\ \hline
AdaBoost (50 trees) & 0.92&6.45&55.27\\ \hline
AdaBoost (200 trees) & 0.94&3.21&202.58\\ \hline
XGBoost (20 trees)& 0.71&13.25&15.65\\ \hline
XGBoost (50 trees)& 0.96&2.59&30.92\\ \hline
XGBoost (200 trees)& 0.97&2.01&105.67\\ \hline
Random Forest (20 trees) & 0.96&1.66&9.88 \\ \hline
Random Forest (50 trees) &0.97&1.44&26.68 \\ \hline
\bf{Extra Trees (20 trees)} & \bf{0.99}&\bf{0.81}&7.25\\ \hline
\bf{Extra Trees (50 trees)} &\bf{0.99}&\bf{0.21}&21.32\\ \hline
\end{tabular}
\end{table}

Algorithms using an ensemble of decision trees have achieved state of the art results for various machine learning tasks~\cite{dietterich2000experimental}. As a non-parametric method like decision trees performed better than parametric methods like linear regression, we decided to explore both boosting and bagging decision trees. Ensemble-based methods have been successful in tackling problems with sequential components~\cite{seker2013ensemble, kumar2006forecasting}. While AdaBoost and XGBoost are tree-based ensemble boosting algorithms in which each successive tree harnesses the decision made by the previous tree, bagged algorithms like Random Forest(RFs) and ERTs make a decision based on the average of many different trees. For both boosting and bagging, weak learners are utilized in the form of trees with limited depth. Boosting models are sequential learners and harnesses weak learners in sequence. As bagged models use many weak tree-based learners in parallel, and hence can be parallelized in the order of the number of processors. As the time of training is essential for a real-time application, we choose bagged decision trees and in particular, ERTs as they outperform RFs for our experiments. Table~\ref{tab:estimators_compare} demonstrates the performance of all the different algorithms trained on the first 200 time steps for predicting the next 300 time steps. 

  ERTs use an ensemble of decision trees in which a node split is selected completely randomly with respect to both variable index and variable splitting value.  ERTs are very good generalized learners and perform better in the presence of noisy features. As compared to RFs, ERTs decrease the variance and increase the bias by randomly selecting a node split independent of the splitting value. Both  RFs and ERTs can utilize bootstrap aggregation wherein each weak learner builds a model based on a random sample of observations from the training data, with replacement. Bootstrap aggregation helps in reducing variance in bagged ensembles.  

\begin{figure*}[!ht]
\centering
\includegraphics[width = 0.85\textwidth, keepaspectratio=true]{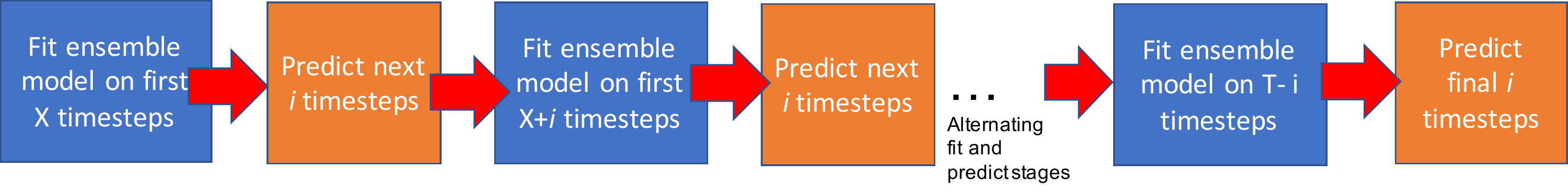}
\caption{The proposed model using ERTs to predict temperature profiles for additive manufacturing processes. It is to be noted that the number of data-points predicted at each step is not the same as the number of data-points for each voxel. This is because the model predicts not only the temperature of the newly created voxels but also the temperature of the same voxels present in the training set at a later time-step. }
\label{fig:iterativeModel}
\end{figure*}

Researchers have proposed rolling recursive or iterative autoregressive moving average modeling~\cite{guerrero1995recursive} for time series prediction. In this work, we decided to explore iterative prediction based on ERTs as we have a combination of historical as well as spatiotemporal features. We propose an iterative model in which an initial model is first developed based on the ground-truth data. Then, the data points predicted by the initial model is added to the ground-truth data to develop a model for the next stage, which predicts the temperature profile of voxels for future time-steps. We iteratively keep predicting future time-steps using predicted temperature profiles from the previous stage alongside ground-truth data. Figure~\ref{fig:iterativeModel} demonstrates the iterative learning process of our proposed model.

\section{Experimental Results}
	\label{sec:results}
In this section, we present the experimental settings and describe the results of the proposed system for predicting temperature profiles in an AM process.

All experiments are carried out using NVIDIA DIGITS DevBox with a Core i7-5930K 6 Core 3.5GHz desktop processor, 64GB DDR4 RAM.  The python VTK librarywas used for processing and converting the voxel data. The data preprocessing, as well as most of the regression models,  were implemented using Scikit-Learn~\cite{pedregosa2011scikit}. The XGBoost package~\cite{chen2015xgboost} was utilized for creating the xgboost model. The ARIMA model was trained from the statsmodels package~\cite{seabold2010statsmodels}.  

For the iterative model, we performed extensive grid-search across various sizes of time step intervals and found the best results when the time step interval was equal to 20. For the experiments, we evaluate with different combinations and ratios of train and test splits. It is to be noted that instead of splitting the train and test set based on a fixed fraction, we divided the dataset based on the timesteps. For instance in Table~\ref{tab:timesteps_combinations}, we use data points up to 1000, 800, 500 and 300  timesteps for training and then we predict the next 200, 400, 700, and 900  timesteps respectively. For instance, when we use 800 timesteps for training and 400 for the test set, it corresponds to about 4.34 million training data points and 4.71 million test data points.

\begin{table*}[]
\centering
\caption{Comparison of combinations of time-steps used for training and test in the iterative model (with corresponding $R^2$ and \% MAE). We vary the number of time-steps used for training and validation. The total number of time-steps - sum of the training and validation time-steps are always equal to 1200.}
\label{tab:timesteps_combinations}
\scalebox{1.0}{
\begin{tabular}{|c|c|c|c|c|c|}
\hline
\multicolumn{2}{|c|}{\textbf{Training}} &
      \multicolumn{2}{c|}{\textbf{Test}} & \multirow{1}{*}{\textbf{$R^2$}}& \multirow{1}{*}{\textbf{\% MAE}}\\
 \textbf{No. of timesteps} & \textbf{No. of datapoints } &\textbf{No. of timesteps} & \textbf{No. of datapoints } & & \\   
 & \textbf{(in millions)} &  & \textbf{(in millions)} & & \\ \hline

1000&6.75&200&2.30&0.992&0.289\\ \hline
800&4.34&400&4.71&0.989&0.679\\ \hline
500&1.72&700&7.33&0.982&1.329\\ \hline
300&0.63&900&8.42&0.972&1.848\\ \hline
\end{tabular}}
\end{table*}
  
\begin{table*}[]
\centering
\caption{Comparison of proposed iterative model with a direct model that directly predicts the temperature of subsequent points. We present the time taken as well as regression metrics (corresponding $R^2$ and \% MAE) for both the models. The initial number of time-steps used for training is set to 200 and the size of the iteration is set as 20 time-steps.  We vary the number of future time-steps predicted.}
\label{tab:timesteps_compare}
\scalebox{1.0}{
\begin{tabular}{|c|c|c|c|c|c|c|c|}
\hline
\multirow{1}{*}{\textbf{Iterations}} & {\textbf{Future}} &
\multicolumn{3}{c|}{\textbf{Iterative Model}} &
      \multicolumn{3}{c|}{\textbf{Standard Model}} \\
&{\textbf{Timesteps}}&&&&&& \\
&{\textbf{Predicted}}&&&&&& \\
 &&Time &\textbf{$R^2$} & \textbf{\% MAE} &Time&\textbf{$R^2$} & \textbf{\% MAE} \\ 
 &&(in seconds) & &   & (in seconds) &  &  \\ \hline
10&200&68.69&0.989&0.675&0.293&0.921&5.39\\ \hline
20&400&137.08&0.978&1.444&0.308&0.906&5.71\\ \hline
30&600&210.04&0.976&1.489&0.317&0.876&6.07\\ \hline
40&800&278.61&0.971&1.903&0.480&0.861&6.55\\ \hline
50&1000&353.96&0.969&1.721&0.590&0.794&6.63\\ \hline
\end{tabular}}
\end{table*}

Table~\ref{tab:timesteps_compare} compares the timing and regression metrics for the proposed iterative model with a standard non-iterative model that directly predicts temperatures of future time steps varying between 200 to 1000. This experimental design of selecting training data based on timesteps instead of layers also helps in generalizing the training set-up. For instance, the first 200 timesteps would represent a few completed layers and an incomplete layer. The same intuition follows for the timesteps in the test set. By training on different timesteps allows us to generalize the framework to different shapes. Although the direct model is much faster, the iterative model performs much better than the direct model. For instance, while predicting the temperature for 1000 future time steps, the iterative model takes 353.96 seconds, the direct model requires 0.29 seconds. However, we can observe that the \% MAE value of the direct model is much worse as compared to the iterative model. While the iterative model has $R^2$ between 0.97 and 0.99 and \% MAE between 0.68 to 1.73 \%, the direct model has $R^2$ between 0.79 and 0.92 and \% MAE between 5.39 to 6.63 \%. 

\begin{table}[]
\centering
\caption{Comparison of $R^2$ and Mean Absolute Error\% across the different types of voxel }
\label{tab:voxels_compare}
\begin{tabular}{|l|l|c|c|}
\hline
\textbf{Type of voxel} & \textbf{\% of overall } &  \textbf{$R^2$} & \textbf{\% MAE}\\
&\textbf{voxels}&&\\ \hline
Interior&40.15&0.990&0.916\\ \hline
Edge (Lateral)&4.92&0.992&0.898\\ \hline
Edge (Longitudinal)&5.09&0.988&0.923\\ \hline
Edge (Vertical)&49.20&0.989&0.918\\ \hline
Edge (Diagonal)&0.63&0.988&0.926\\ \hline
\end{tabular}
\end{table}

\begin{table}[]
\centering
\caption{Comparison of number of trees/estimators in the ensemble. As we vary the number of estimators, we present the trade-off in the form of time and $R^2$ and Mean Absolute Error\%. The number of voxels predicted in each iteration is 25, and there are 40 steps in each iteration }
\label{tab:trees_compare}
\scalebox{1.0}{
\begin{tabular}{|l|l|c|c|}
\hline
\textbf{No. of estimators} & \textbf{Overall Time } &  \textbf{$R^2$} & \textbf{\% MAE}\\ 
& (in seconds)& & \\ \hline
4&154.5&0.964&2.14\\ \hline
10&257.5&0.970&1.38\\ \hline
20&493.2&0.975&1.29\\ \hline
50&902.4&0.981&1.03\\ \hline
\end{tabular}}
\end{table}

\begin{figure*}[!h]
\centering
\includegraphics[width=0.85\textwidth]{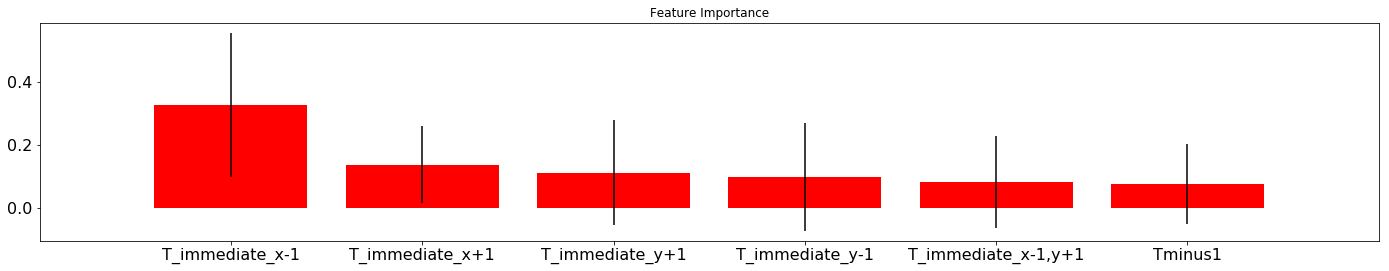}
\caption{The feature importance for the top input features in the ensemble iterative approach}
\label{fig:featureImportance}
\end{figure*}

\begin{figure*}[!ht]
\centering
\subfigure[4 trees]{%
\includegraphics[width=.43\textwidth]{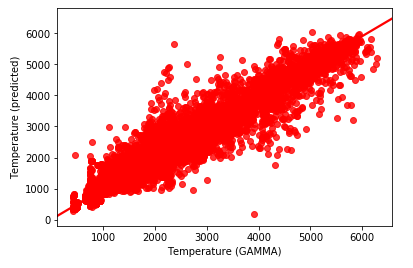}
}
\quad
\subfigure[10 trees]{%
\includegraphics[width=.43\textwidth]{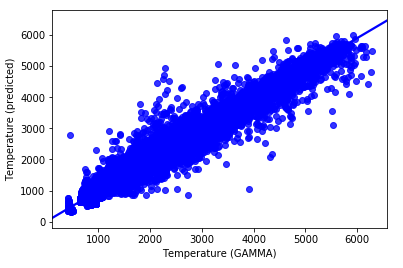}
}
\quad
\subfigure[20 trees]{%
\includegraphics[width=.43\textwidth]{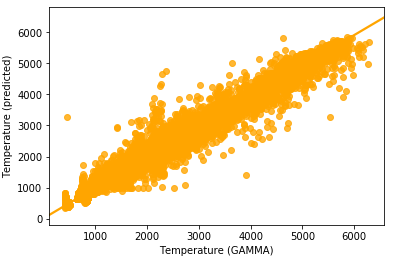}
}
\quad
\subfigure[50 trees]{%
\includegraphics[width=.43\textwidth]{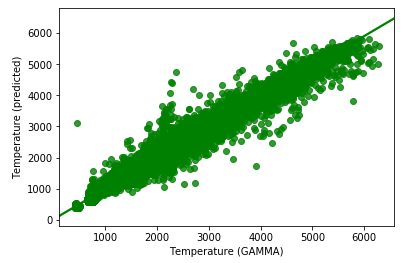}
}
\caption{Scatterplot for predicted vs. FEM temperatures. As the number of estimators/trees increase, the prediction accuracy improves. }
\label{fig:scatter}
\end{figure*}

The results in Table~\ref{tab:voxels_compare} illustrates that interior and edge (vertical) voxels comprise the bulk of the voxels (40.15\% and 49.20\%). This is anticipated as for any new layer created, none of the voxels in the new layer would have a vertical neighbor until a new layer is deposited. We also find that there is no significant difference in the prediction accuracy between the type of voxels. This demonstrates further that our iterative prediction model is able to learn the temperature profiles for both edge voxels as well as interior voxels.  

Table~\ref{tab:trees_compare} depict how varying the number of estimators (trees, in the case of ERTs) impacts the overall time (sum of the training and prediction times). As expected, the \% MAE reduces and $R^2$ increases as the number of estimators increase. The variance of bagged ensembles reduce as more trees are used to make the prediction, and MAE reduces with variance. However, as the overall time increases with the number of estimators, any deployed system would need to make a trade-off between reducing the \% MAE and the cost and time of the available computing resources.

Figure~\ref{fig:featureImportance} illustrates the impact of the temperature profiles of the voxels immediately surrounding the target voxel for which we are predicting the temperature profile.  The voxels on the $x$-axis have a more significant impact than the voxels on the $y$-axis. This is expected as the direction of the laser is towards the $x$-axis. Further, the importance of the ${T\_{immediate}\_{(y+1)}}$ and ${T\_{immediate}\_{(y-1)}}$ features are equal and this is also unsurprising as the laser path zig-zags on the $y$-axis during the AM process (as illustrated in ~\ref{fig:additive}b) and is therefore agnostic of the directionality in the $y$-axis. Figure~\ref{fig:scatter} depicts the scatterplot for the predicted vs. the ground-truth FEM voxel temperatures. We can observe that the prediction accuracy increases with the number of estimators. Further, we have fewer outliers when the number of estimators is higher. This is expected as bagged ensembles perform well based on crowd-sourcing the prediction of weak learners which are likely to have a high bias on their own but have low bias overall as an ensemble. 

The primary motivation of this work was to develop an ML-aided framework that can reduce or replace FEM simulations. Hence, it was very important to have a model that has a low MAE guarantee. ERTs are especially effective at creating data-driven rules for handling different kinds of data points. For a voxel that has been created long ago, such as in the first layer, the temperature of the voxel would not change as a new voxel is created at the topmost layer. However, the temperature of a given voxel created few time steps or a voxel created many time steps before but immediately below a newly created voxel would be high. Not only are ERTs fast to train, but they are also easy to interpret as we can rank the features as well as visualize the different candidate trees. Interpretability of algorithms is extremely important in the scientific and engineering community.

\section{Conclusions and Future Work}
	\label{sec:conclusion}
This paper presents essential components of a scientific framework for a model-based real-time AM control system.  The proposed approach utilizes extremely randomized trees - an ensemble of bagged decision trees as the regression algorithm iteratively using temperatures of prior voxels and laser information as inputs to predict temperatures of subsequent voxels and is able to achieve \% MAE below 1\% for predicting temperature profiles.  One of the advantages of a real-time system is instead of training a prior model ahead of time, one can be trained in-situ. It is crucial for the versatility of the AM ML-driven simulation process, especially as factors such as laser path, laser speed, and laser temperature can largely influence the temperature profile.

 In the future, we plan to explore the impact of voxel mesh size on the prediction results across coarse to finer mesh. The next goal of this framework is to be part of an interleaved FEM-ML control system that harnesses the temperature profile of the odd layer (Layer $i$) calculated using FEM to predict the subsequent even layer (Layer $i+1$). Layer $i+2$ will then be calculated using FEM simulation, and Layer $i+3$ will be predicted. This can accelerate the speed of simulations by nearly a factor of two, hopefully without impacting the accuracy significantly. Although this work restricts itself to temperature profile prediction for an AM process, the same idea can be extended to related manufacturing processes such as incremental forming~\cite{kurra2015modeling}. In general, this work can be extended to any phenomenon which utilizes partial differential equation based modeling.

\section*{Acknowledgment}
This work was performed under the following financial assistance award 70NANB19H005 from U.S. Department of Commerce, National Institute of Standards and Technology as part of the Center for Hierarchical Materials Design (CHiMaD). Partial support is also acknowledged from DOE awards DE-SC0014330, DE-SC0019358.


\bibliographystyle{unsrt}
\bibliography{additive_bib}

\begin{thebibliography}{10}

\bibitem{watson2018decision}
JK~Watson and KMB Taminger.
\newblock A decision-support model for selecting additive manufacturing versus
  subtractive manufacturing based on energy consumption.
\newblock {\em Journal of Cleaner Production}, 176:1316--1322, 2018.

\bibitem{ding2016development}
Yaoyu Ding, James Warton, and Radovan Kovacevic.
\newblock Development of sensing and control system for robotized laser-based
  direct metal addition system.
\newblock {\em Additive Manufacturing}, 10:24--35, 2016.

\bibitem{ding2011thermo}
J~Ding, P~Colegrove, Jorn Mehnen, Supriyo Ganguly, PM~Sequeira Almeida, F~Wang,
  and S~Williams.
\newblock Thermo-mechanical analysis of wire and arc additive layer
  manufacturing process on large multi-layer parts.
\newblock {\em Computational Materials Science}, 50(12):3315--3322, 2011.

\bibitem{yan2018data}
Wentao Yan, Stephen Lin, Orion~L Kafka, Yanping Lian, Cheng Yu, Zeliang Liu,
  Jinhui Yan, Sarah Wolff, Hao Wu, Ebot Ndip-Agbor, et~al.
\newblock Data-driven multi-scale multi-physics models to derive
  process--structure--property relationships for additive manufacturing.
\newblock {\em Computational Mechanics}, 61(5):521--541, 2018.

\bibitem{correa2018new}
Jorge~E Correa, Ricardo Toro, and Placid~M Ferreira.
\newblock A new paradigm for organizing networks of computer numerical control
  manufacturing resources in cloud manufacturing.
\newblock {\em Procedia Manufacturing}, 26:1318--1329, 2018.

\bibitem{garcia2019sustainable}
Daniel~J Garcia, Mojtaba Mozaffar, Huaqing Ren, Jorge~E Correa, Kornel Ehmann,
  Jian Cao, and Fengqi You.
\newblock Sustainable manufacturing with cyber-physical discrete manufacturing
  networks: Overview and modeling framework.
\newblock {\em Journal of Manufacturing Science and Engineering},
  141(2):021013, 2019.

\bibitem{smith2016thermodynamically}
Jacob Smith, Wei Xiong, Jian Cao, and Wing~Kam Liu.
\newblock Thermodynamically consistent microstructure prediction of additively
  manufactured materials.
\newblock {\em Computational mechanics}, 57(3):359--370, 2016.

\bibitem{mozaffar2019acceleration}
Mojtaba Mozaffar, Ebot Ndip-Agbor, Stephen Lin, Gregory~J Wagner, Kornel
  Ehmann, and Jian Cao.
\newblock Acceleration strategies for explicit finite element analysis of metal
  powder-based additive manufacturing processes using graphical processing
  units.
\newblock {\em Computational Mechanics}, pages 1--16, 2019.

\bibitem{li2018residual}
C~Li, ZY~Liu, XY~Fang, and YB~Guo.
\newblock Residual stress in metal additive manufacturing.
\newblock {\em Procedia Cirp}, 71:348--353, 2018.

\bibitem{geurts2006extremely}
Pierre Geurts, Damien Ernst, and Louis Wehenkel.
\newblock Extremely randomized trees.
\newblock {\em Machine learning}, 63(1):3--42, 2006.

\bibitem{frazier2014metal}
William~E Frazier.
\newblock Metal additive manufacturing: a review.
\newblock {\em Journal of Materials Engineering and Performance},
  23(6):1917--1928, 2014.

\bibitem{wong2012review}
Kaufui~V Wong and Aldo Hernandez.
\newblock A review of additive manufacturing.
\newblock {\em ISRN Mechanical Engineering}, 2012, 2012.

\bibitem{ngo2018additive}
Tuan~D Ngo, Alireza Kashani, Gabriele Imbalzano, Kate~TQ Nguyen, and David Hui.
\newblock Additive manufacturing (3d printing): A review of materials, methods,
  applications and challenges.
\newblock {\em Composites Part B: Engineering}, 2018.

\bibitem{mueller2012additive}
Bernhard Mueller.
\newblock Additive manufacturing technologies--rapid prototyping to direct
  digital manufacturing.
\newblock {\em Assembly Automation}, 32(2), 2012.

\bibitem{kruth1998progress}
J-P Kruth, Ming-Chuan Leu, and Terunaga Nakagawa.
\newblock Progress in additive manufacturing and rapid prototyping.
\newblock {\em Cirp Annals}, 47(2):525--540, 1998.

\bibitem{faludi2015comparing}
Jeremy Faludi, Cindy Bayley, Suraj Bhogal, and Myles Iribarne.
\newblock Comparing environmental impacts of additive manufacturing vs
  traditional machining via life-cycle assessment.
\newblock {\em Rapid Prototyping Journal}, 21(1):14--33, 2015.

\bibitem{matilainen2014characterization}
Ville Matilainen, Heidi Piili, Antti Salminen, Tatu Syv{\"a}nen, and Olli
  Nyrhil{\"a}.
\newblock Characterization of process efficiency improvement in laser additive
  manufacturing.
\newblock {\em Physics Procedia}, 56:317--326, 2014.

\bibitem{huang2013additive}
Samuel~H Huang, Peng Liu, Abhiram Mokasdar, and Liang Hou.
\newblock Additive manufacturing and its societal impact: a literature review.
\newblock {\em The International Journal of Advanced Manufacturing Technology},
  67(5-8):1191--1203, 2013.

\bibitem{peyre2008analytical}
P~Peyre, P~Aubry, R~Fabbro, R~Neveu, and Arnaud Longuet.
\newblock Analytical and numerical modelling of the direct metal deposition
  laser process.
\newblock {\em Journal of Physics D: Applied Physics}, 41(2):025403, 2008.

\bibitem{qi2006numerical}
Huan Qi, Jyotirmoy Mazumder, and Hyungson Ki.
\newblock Numerical simulation of heat transfer and fluid flow in coaxial laser
  cladding process for direct metal deposition.
\newblock {\em Journal of applied physics}, 100(2):024903, 2006.

\bibitem{fish2007first}
Jacob Fish and Ted Belytschko.
\newblock {\em A first course in finite elements}, volume~1.
\newblock John Wiley \& Sons New York, 2007.

\bibitem{stender2018thermal}
Michael~E Stender, Lauren~L Beghini, Joshua~D Sugar, Michael~G Veilleux,
  Samuel~R Subia, Thale~R Smith, Christopher~W San~Marchi, Arthur~A Brown, and
  Daryl~J Dagel.
\newblock A thermal-mechanical finite element workflow for directed energy
  deposition additive manufacturing process modeling.
\newblock {\em Additive Manufacturing}, 21:556--566, 2018.

\bibitem{smith2016linking}
Jacob Smith, Wei Xiong, Wentao Yan, Stephen Lin, Puikei Cheng, Orion~L Kafka,
  Gregory~J Wagner, Jian Cao, and Wing~Kam Liu.
\newblock Linking process, structure, property, and performance for metal-based
  additive manufacturing: computational approaches with experimental support.
\newblock {\em Computational Mechanics}, 57(4):583--610, 2016.

\bibitem{wolff2017framework}
Sarah~J Wolff, Stephen Lin, Eric~J Faierson, Wing~Kam Liu, Gregory~J Wagner,
  and Jian Cao.
\newblock A framework to link localized cooling and properties of directed
  energy deposition (ded)-processed ti-6al-4v.
\newblock {\em Acta Materialia}, 132:106--117, 2017.

\bibitem{francois2017modeling}
Marianne~M Francois, Amy Sun, Wayne~E King, Neil~Jon Henson, Damien Tourret,
  Ccut~Allan Bronkhorst, Neil~N Carlson, Christopher~Kyle Newman, Terry~Scot
  Haut, Jozsef Bakosi, et~al.
\newblock Modeling of additive manufacturing processes for metals: Challenges
  and opportunities.
\newblock {\em Current Opinion in Solid State and Materials Science},
  21(LA-UR-16-24513), 2017.

\bibitem{kalidindi2015data}
Surya~R Kalidindi.
\newblock Data science and cyberinfrastructure: critical enablers for
  accelerated development of hierarchical materials.
\newblock {\em International Materials Reviews}, 60(3):150--168, 2015.

\bibitem{olson1997computational}
Gregory~B Olson.
\newblock Computational design of hierarchically structured materials.
\newblock {\em Science}, 277(5330):1237--1242, 1997.

\bibitem{agrawal2019deep}
Ankit Agrawal and Alok Choudhary.
\newblock Deep materials informatics: Applications of deep learning in
  materials science.
\newblock {\em MRS Communications}, pages 1--14, 2019.

\bibitem{sundararaghavan2007linear}
Veera Sundararaghavan and Nicholas Zabaras.
\newblock Linear analysis of texture--property relationships using
  process-based representations of rodrigues space.
\newblock {\em Acta Materialia}, 55(5):1573--1587, 2007.

\bibitem{cang2017microstructure}
Ruijin Cang, Yaopengxiao Xu, Shaohua Chen, Yongming Liu, Yang Jiao, and Max~Yi
  Ren.
\newblock Microstructure representation and reconstruction of heterogeneous
  materials via deep belief network for computational material design.
\newblock {\em Journal of Mechanical Design}, 139(7):071404, 2017.

\bibitem{paul2019opv}
Arindam Paul, Alona Furmanchuk, Wei-keng Liao, Alok Choudhary, and Ankit
  Agrawal.
\newblock Property prediction of organic donor molecules for photovoltaic
  applications using extremely randomized trees.
\newblock {\em Molecular Informatics}, 2019.

\bibitem{kusne2014fly}
Aaron~Gilad Kusne, Tieren Gao, Apurva Mehta, Liqin Ke, Manh~Cuong Nguyen,
  Kai-Ming Ho, Vladimir Antropov, Cai-Zhuang Wang, Matthew~J Kramer, Christian
  Long, et~al.
\newblock On-the-fly machine-learning for high-throughput experiments: search
  for rare-earth-free permanent magnets.
\newblock {\em Scientific reports}, 4, 2014.

\bibitem{paul2019transfer}
Arindam Paul, Dipendra Jha, Reda Al-Bahrani, Wei-keng Liao, Alok Choudhary, and
  Ankit Agrawal.
\newblock Transfer learning using ensemble neural networks for organic solar
  cell screening.
\newblock In {\em 2019 International Joint Conference on Neural Networks
  (IJCNN)}. IEEE, 2019.

\bibitem{jha2019irnet}
Dipendra Jha, Logan Ward, Zijiang Yang, Christopher Wolverton, Ian Foster,
  Wei-keng Liao, Alok Choudhary, and Ankit Agrawal.
\newblock Irnet: A general purpose deep residual regression framework for
  materials discovery.
\newblock In {\em Proceedings of the 25th ACM SIGKDD International Conference
  on Knowledge Discovery \& Data Mining}. ACM, 2019.

\bibitem{jung2019efficient}
Jaimyun Jung, Jae~Ik Yoon, Hyung~Keun Park, Jin~You Kim, and Hyoung~Seop Kim.
\newblock An efficient machine learning approach to establish
  structure-property linkages.
\newblock {\em Computational Materials Science}, 156:17--25, 2019.

\bibitem{jha2018elemnet}
Dipendra Jha, Logan Ward, Arindam Paul, Wei-keng Liao, Alok Choudhary, Chris
  Wolverton, and Ankit Agrawal.
\newblock Elemnet: Deep learning the chemistry of materials from only elemental
  composition.
\newblock {\em Scientific reports}, 8(1):17593, 2018.

\bibitem{rahnama2018machine}
Alireza Rahnama, Sam Clark, and Seetharaman Sridhar.
\newblock Machine learning for predicting occurrence of interphase
  precipitation in hsla steels.
\newblock {\em Computational Materials Science}, 154:169--177, 2018.

\bibitem{paul2018chemixnet}
Arindam Paul, Dipendra Jha, Reda Al-Bahrani, Wei-keng Liao, Alok Choudhary, and
  Ankit Agrawal.
\newblock Chemixnet: Mixed dnn architectures for predicting chemical properties
  using multiple molecular representations.
\newblock In {\em Proceedings of the Workshop on Molecules and Materials at the
  32nd Conference on Neural Information Processing Systems}, 2018.

\bibitem{pozun2012optimizing}
Zachary~D Pozun, Katja Hansen, Daniel Sheppard, Matthias Rupp, Klaus-Robert
  M{\"u}ller, and Graeme Henkelman.
\newblock Optimizing transition states via kernel-based machine learning.
\newblock {\em The Journal of chemical physics}, 136(17):174101, 2012.

\bibitem{paul2018sampling}
Arindam Paul, Pinar Acar, Ruoqian Liu, Wei-keng Liao, Alok Choudhary, Veera
  Sundararaghavan, and Ankit Agrawal.
\newblock Data sampling schemes for microstructure design with vibrational
  tuning constraints.
\newblock {\em AIAA Journal}, 56(3):1239--1250, 2018.

\bibitem{ramprasad2017npjreview}
Rampi Ramprasad, Rohit Batra, Ghanshyam Pilania, Arun Mannodi-Kanakkithodi, and
  Chiho Kim.
\newblock Machine learning in materials informatics: recent applications and
  prospects.
\newblock {\em npj Computational Materials}, 3(1):54, 2017.

\bibitem{paul2019feedback}
Arindam Paul, Pinar Acar, Wei-keng Liao, Alok Choudhary, Veera Sundararaghavan,
  and Ankit Agrawal.
\newblock Microstructure optimization with constrained design objectives using
  machine learning-based feedback-aware data-generation.
\newblock {\em Computational Materials Science}, 160:334–351, 2019.

\bibitem{jha2019padnet}
Dipendra Jha, Aaron~Gilad Kusne, Nam Nguyen, Wei-keng Liao, Alok Choudhary, and
  Ankit Agrawal.
\newblock Peak area detection network for directly learning phase regions from
  raw x-ray diffraction patterns.
\newblock In {\em 2019 International Joint Conference on Neural Networks
  (IJCNN)}. IEEE, 2019.

\bibitem{mozaffar2018data}
Mojtaba Mozaffar, Arindam Paul, Reda Al-Bahrani, Sarah Wolff, Alok~Nidhi
  Choudhary, Ankit Agrawal, Kornel Ehmann, and Jian Cao.
\newblock Data-driven prediction of the high-dimensional thermal history in
  directed energy deposition processes via recurrent neural networks.
\newblock {\em Manufacturing Letters}, 18:35--39, 2018.

\bibitem{baturynska2018optimization}
Ivanna Baturynska, Oleksandr Semeniuta, and Kristian Martinsen.
\newblock Optimization of process parameters for powder bed fusion additive
  manufacturing by combination of machine learning and finite element method: A
  conceptual framework.
\newblock {\em Procedia CIRP}, 67:227--232, 2018.

\bibitem{choy20163d}
Christopher~B Choy, Danfei Xu, JunYoung Gwak, Kevin Chen, and Silvio Savarese.
\newblock 3d-r2n2: A unified approach for single and multi-view 3d object
  reconstruction.
\newblock In {\em European conference on computer vision}, pages 628--644.
  Springer, 2016.

\bibitem{scime2018anomaly}
Luke Scime and Jack Beuth.
\newblock Anomaly detection and classification in a laser powder bed additive
  manufacturing process using a trained computer vision algorithm.
\newblock {\em Additive Manufacturing}, 19:114--126, 2018.

\bibitem{scime2018multi}
Luke Scime and Jack Beuth.
\newblock A multi-scale convolutional neural network for autonomous anomaly
  detection and classification in a laser powder bed fusion additive
  manufacturing process.
\newblock {\em Additive Manufacturing}, 24:273--286, 2018.

\bibitem{scime2019using}
Luke Scime and Jack Beuth.
\newblock Using machine learning to identify in-situ melt pool signatures
  indicative of flaw formation in a laser powder bed fusion additive
  manufacturing process.
\newblock {\em Additive Manufacturing}, 25:151--165, 2019.

\bibitem{tapia2014review}
Gustavo Tapia and Alaa Elwany.
\newblock A review on process monitoring and control in metal-based additive
  manufacturing.
\newblock {\em Journal of Manufacturing Science and Engineering},
  136(6):060801, 2014.

\bibitem{ulrich2005timeseries}
Ulrich Helfenstein.
\newblock {\em ARMA and ARIMA Models}.
\newblock American Cancer Society, 2005.

\bibitem{dietterich2000experimental}
Thomas~G Dietterich.
\newblock An experimental comparison of three methods for constructing
  ensembles of decision trees: Bagging, boosting, and randomization.
\newblock {\em Machine learning}, 40(2):139--157, 2000.

\bibitem{seker2013ensemble}
Sadi~Evren Seker, Cihan Mert, Khaled Al-Naami, Ugur Ayan, and Nuri Ozalp.
\newblock Ensemble classification over stock market time series and economy
  news.
\newblock In {\em 2013 IEEE International Conference on Intelligence and
  Security Informatics}, pages 272--273. IEEE, 2013.

\bibitem{kumar2006forecasting}
Manish Kumar and M~Thenmozhi.
\newblock Forecasting stock index movement: A comparison of support vector
  machines and random forest.
\newblock In {\em Indian institute of capital markets 9th capital markets
  conference paper}, 2006.

\bibitem{guerrero1995recursive}
V{\'\i}ctor~M Guerrero and J~Mart{\'\i}nez.
\newblock A recursive arima-based procedure for disaggregating a time series
  variable using concurrent data.
\newblock {\em Test}, 4(2):359--376, 1995.

\bibitem{pedregosa2011scikit}
Fabian Pedregosa, Ga{\"e}l Varoquaux, Alexandre Gramfort, Vincent Michel,
  Bertrand Thirion, Olivier Grisel, Mathieu Blondel, Peter Prettenhofer, Ron
  Weiss, Vincent Dubourg, et~al.
\newblock Scikit-learn: Machine learning in python.
\newblock {\em Journal of machine learning research}, 12(Oct):2825--2830, 2011.

\bibitem{chen2015xgboost}
Tianqi Chen, Tong He, Michael Benesty, Vadim Khotilovich, and Yuan Tang.
\newblock Xgboost: extreme gradient boosting.
\newblock {\em R package version 0.4-2}, pages 1--4, 2015.

\bibitem{seabold2010statsmodels}
Skipper Seabold and Josef Perktold.
\newblock Statsmodels: Econometric and statistical modeling with python.
\newblock In {\em Proceedings of the 9th Python in Science Conference},
  volume~57, page~61. Scipy, 2010.

\bibitem{kurra2015modeling}
Suresh Kurra, Nasih~Hifzur Rahman, Srinivasa~Prakash Regalla, and Amit~Kumar
  Gupta.
\newblock Modeling and optimization of surface roughness in single point
  incremental forming process.
\newblock {\em Journal of Materials Research and Technology}, 4(3):304--313,
  2015.

\end{thebibliography}

\end{document}